\pdfoutput=1

\documentclass[11pt]{article}

\usepackage[]{acl}

\usepackage{times}
\usepackage{latexsym}

\usepackage[T1]{fontenc}

\usepackage[utf8]{inputenc}

\usepackage{microtype}

\usepackage{graphicx}
\usepackage{amsmath}
\usepackage{subcaption}
\graphicspath{{figures/}}
\usepackage{dsfont}
\usepackage{xcolor}
\usepackage{amssymb}
\usepackage{bm}
\usepackage{booktabs}

\newcommand{\cut}[1]{}

\title{HuCurl: Human-induced Curriculum Discovery}

\author{Mohamed Elgaar \and Hadi Amiri \\
  Department of Computer Science \\
  University of Massachusetts Lowell\\
  \texttt{\{melgaar,hadi\}@cs.uml.edu}}

\def\bx{{\mathbf x}}

\def\bpsi{{\mathbf\psi}} 

\def\R{{\mathbf R}}

\begin{document}
\maketitle
\begin{abstract}
We introduce the problem of {\em curriculum discovery} and describe a curriculum learning framework capable of discovering effective curricula in a curriculum space based on prior knowledge about sample difficulty. 
Using annotation entropy and loss as measures of difficulty, we show that
(i): the top-performing discovered curricula for a given model and dataset are often {\em non-monotonic} as opposed to {\em monotonic} curricula in existing literature,
(ii): the prevailing easy-to-hard or hard-to-easy transition curricula are often at the risk of underperforming, and
(iii): the curricula discovered for smaller datasets and models perform well on larger datasets and models respectively. 
The proposed framework encompasses some of the existing curriculum learning approaches and can discover curricula that outperform them across several NLP tasks.



\end{abstract}

\section{Introduction}
Annotation information has been extensively used by previous research in NLP to devise strategies for 
further data collection~\cite{yang2019predicting,dligach2010annotate}, 
model improvement and annotation analysis~\cite{zaidan2008modeling,paun-etal-2018-comparing},
pruning and weighting samples for better learning~\cite{yang2019predicting}, or
efficient use of monetary funds~\cite{dligach2010annotate}. 
Recent studies show consistent positive correlation between difficulty of samples to the model and their level of human agreement~\cite{nie2020can,zaidan2008modeling,yang2019predicting}. Building on these findings, we aim to utilize such prior knowledge about sample difficulty to develop a curriculum learning (CL) framework that is capable of discovering effective curricula for NLP tasks. 

A curriculum is a planned sequence of learning materials and an effective one can improve training of NLP systems~\cite{settles2016trainable,amiri-etal-2017-repeat,zhang2019curriculum,Lalor2020-mz,xu2020curriculum,kreutzer-etal-2021-bandits-dont,agrawal-carpuat-2022-imitation,maharana-bansal-2022-curriculum}. 
CL seeks to improve model generalizability by ordering samples for training based on their latent difficulty~\cite{Bengio2009}. 
Recent work reported efficiency and effectiveness gains through CL~\cite{jiang2018mentornet,castells2020superloss,zhou2020curriculum}, especially in cases of harder tasks and limited or noisy data~\citep{wu2021when}.

Existing CL approaches are designed to learn a {\em single} curriculum that works best for a given model and dataset. However, effective training could be achieved in multiple ways. In addition, existing approaches quantify sample difficulty through model behavior {\em during} training. Although efficient and effective, model behavior can be affected by initialization and training dynamics~\citep{erhan2010does,wu2021when}, which limits the curriculum space that can be examined for finding effective curricula.

This paper advocates a re-imagining of CL paradigms by introducing and formalizing the task of {\em curriculum discovery}, which aims to find effective curricula for a given model and dataset over a curriculum space. The present work specifically focuses on determining {\em when} and in {\em which difficulty order} text data samples should be learned for effective training of NLP systems. We propose a framework that employs prior knowledge about sample difficulty, such as entropy in human annotations, to 
inform an effective and flexible sample weighting scheme for curriculum discovery. 
The framework is capable of discovering optimal curricula (within the space of its weight functions) for any given model and dataset by optimizing 
the weight functions 
and adjusting the difficulty group of data samples as training progresses. The discovered curricula provide useful insights about datasets and models, such as the relative importance of different groups of samples for models or knowledge dependency among samples. 
We illustrate that the proposed framework has the potential to encompass some of the existing CL approaches.



Experimental results show that 
(a): the top-performing discovered curricula for the same model and dataset can be fundamentally dissimilar in their training strategies, 
indicating that effective training can be achieved in multiple ways;
(b): the discovered curricula are often non-monotonic and greatly differ from the known strategies reported in existing literature, indicating that existing curricula, including easy-to-hard transition curricula, are at the risk of underperforming; and 
(c): the curricula discovered on small datasets and models perform exceptionally well on larger datasets and models respectively, illustrating the transferability of the discovered curricula.
The paper presents a new curriculum learning approach that unlike existing approaches can discover multiple high-performing (and often diverse) curricula for each given NLP model and dataset, provide interpretable curricula in terms of sample difficulty, and encompass some of the existing curriculum learning approaches.\footnote{Code and data are available at \url{https://clu.cs.uml.edu/tools/curriculum_discovery.html}.}

\section{Related Work}
Existing CL approaches are designed to learn a {\em single} curriculum that works best for a given model and dataset. They estimate sample difficulty through model behavior during training, quantified by the 
instantaneous loss~\citep{xu2020curriculum,wu2021when}, 
consistency in instantaneous loss~\citep{xu2020curriculum}, 
moving average of loss~\citep{jiang2018mentornet,zhou2020curriculum}, 
transformations of loss~\citep{amiri-etal-2017-repeat,castells2020superloss, chen2021curriculum, vakil-amiri-2022-generic}, 
loss regularization~\citep{kumar2010self,jiang2015self,castells2020superloss}, or 
learnable per-sample confidence~\citep{shu-etal-2021-gu,SaxenaApple,jiang2018mentornet}.
In terms of data ordering,  
sub-sampling approaches sample the easiest or hardest instances at every training iteration~\citep{Bengio2009,kumar2010self,guo2018curriculumnet,platanios2019competence,xu2020curriculum}, 
sample weighting techniques weight instances according to their estimated difficulty~\citep{kumar2010self, jiang2015self,jiang2018mentornet, yang2019predicting,castells2020superloss,zhou2020curriculum}, and 
sample pruning techniques filter hard or noisy instances from data prior to training~\citep{northcutt2021confident}. 
Sub-sampling methods can be cumulative, exclusive or a combination of both. Cumulative approaches add new samples to the ones that have been previously used for training~\citep{guo2018curriculumnet,xu2020curriculum}, while exclusive approaches create a new subset of the data at every training stage~\citep{Bengio2009, zhou2018minimax}. 
In addition, previous research has developed model-driven~\citep{karras2018progressive, morerio2017curriculum,sinha2020curriculum} and task-driven~\citep{caubriere2019curriculum,florensa2017reverse,sarafianos2017curriculum} techniques.

\begin{figure}
    \centering
    \includegraphics[width=\linewidth]{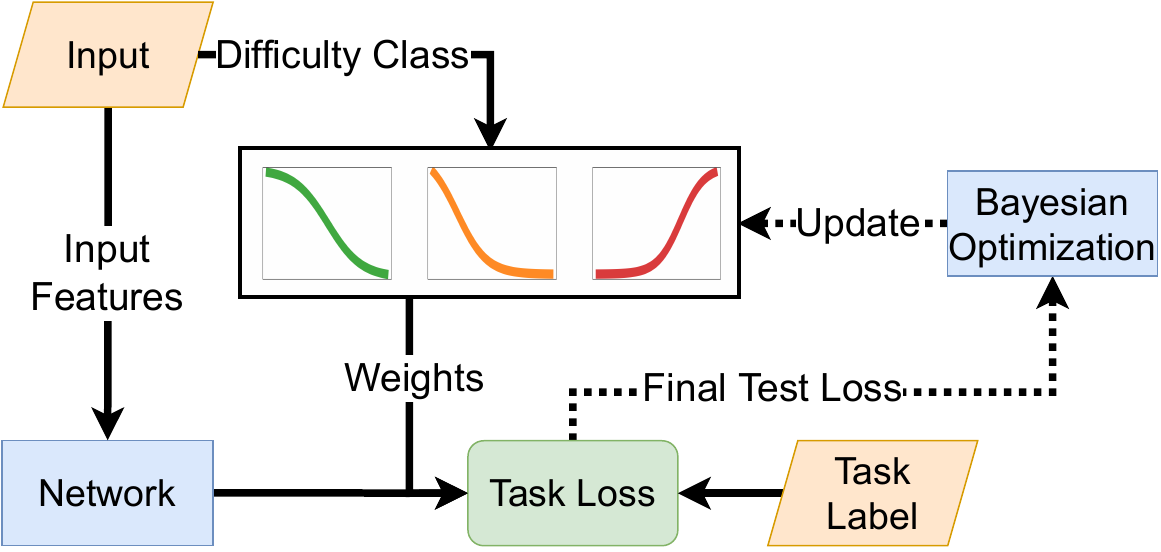}
    \caption{
    The model defines a difficulty score based on prior knowledge about sample difficulty and assigns samples to $k$ difficulty groups before training, e.g., {\it easy}, {\it medium}, and {\it hard} for $k=3$. 
    A curriculum is defined for each difficulty group, which dynamically weights sample losses according to their difficulty groups.
    Each curriculum is defined by a pair of parameters $(r, s)$ that will be optimized to discover an optimized curriculum based on sample difficulty and model behavior.
    }
    \label{fig:model}
\end{figure}

\section{Curriculum Discovery Framework}


We consider the training dataset $\mathcal{D} = \{(\bx_1, y_1), \dots ,(\bx_n, y_n)\}$ of size $n$, where $\bx_i$ denotes the $i$th training sample with the ground-truth label $y_i$ and $\boldsymbol\psi\in [0,1]^{n}$ indicates the initial difficulty estimates of training samples, see \S\ref{sec:scoring}.
The data is initially clustered into $k$ groups of increasing difficulty, e.g. \{\textit{easy}, \textit{medium}, \textit{hard}\} groups for $k=3$, which can be achieved using difficulty score percentiles
or 1-dimensional K-means applied to $\boldsymbol\bpsi$. 
As Figure~\ref{fig:model} shows, the framework develops a separate parameterized weight function for each difficulty group (\S\ref{sec:weighting}), and dynamically weights training samples and adjust their difficulty groups according to the training progress of the downstream model (\S\ref{sec:dyn_class}). Specifically, at training iteration $t$, the weighted loss $\hat{l}_i$ for sample $i$ of the difficulty group $c \in \{1,\dots, k\}$ will be computed as follows:
\begin{equation}
\hat{l}_i = w(t; r_c, s_c) \times l_i,
\label{eq:weight}
\end{equation}
where $l_i$ is the instantaneous loss of sample $i$, and $w(t; r_c, s_c)$ is the weight of sample $i$ in its difficulty group $c$ at training iteration $t$, with class-specific weight function parameters $r_c$ and $s_c$ (see below).

\subsection{Monotonic Curricula}\label{sec:weighting}
\begin{figure}[t]
     \centering
     \begin{subfigure}[htb]{0.6\linewidth}
         \centering
         \includegraphics[width=\linewidth]{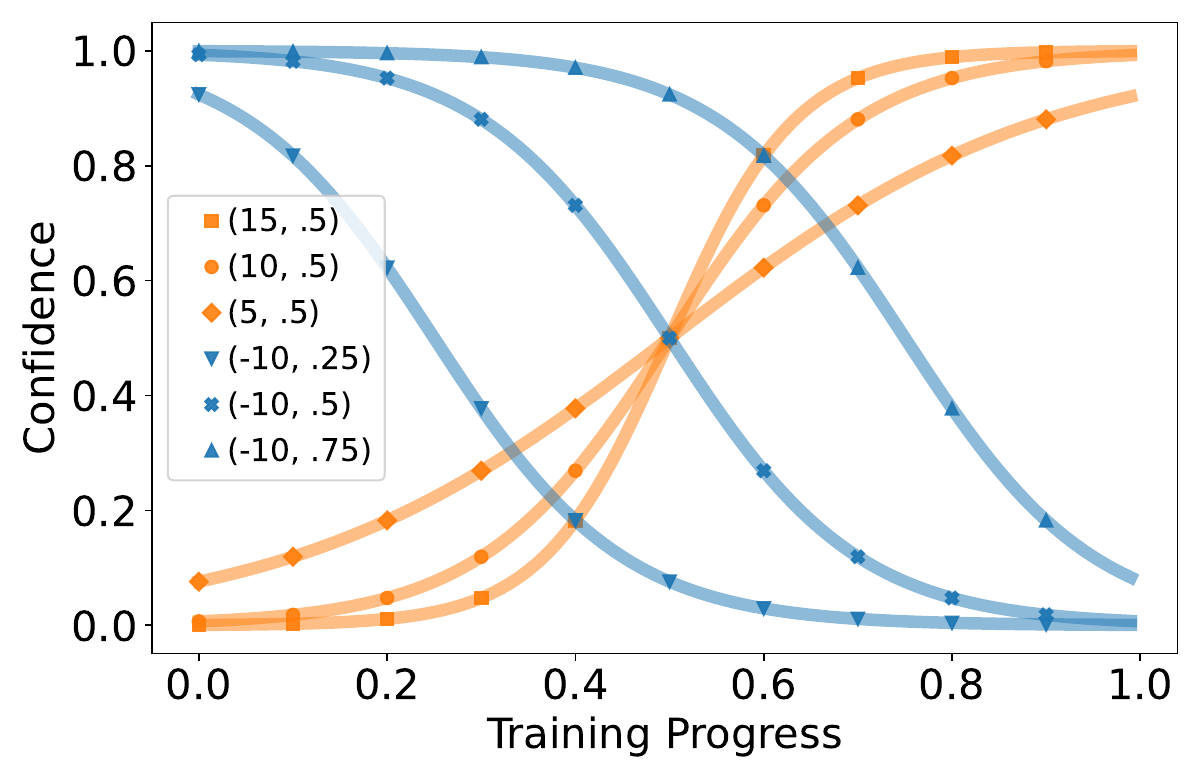}
         \caption{Effect of rate/shift parameters.}
         \label{fig:base_sigmoid}
     \end{subfigure}
     \quad
     \begin{subfigure}[htb]{0.6\linewidth}
         \centering
         \includegraphics[width=\linewidth]{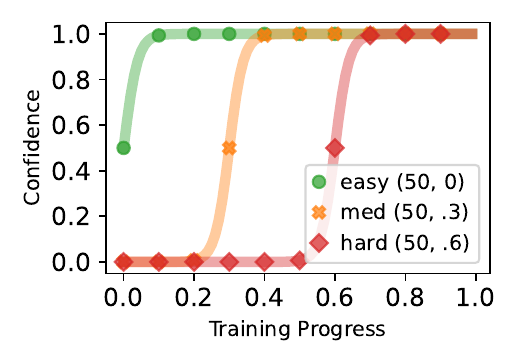}
         \caption{Easy to Hard Curriculum.}
         \label{fig:inc_cfg}
     \end{subfigure}
     \caption{Generalized logistic functions for curriculum discovery. (a) shows the effect of the \textit{rate} and \textit{shift} parameters, $(r,s)$ in (\ref{eq:glf}), shown in the legend respectively. (b) is a specific parameter configuration for a curriculum that first introduces easier samples to a model, and then medium and hard samples as training progresses.}
    \label{fig:logistic_plots}
    \vspace{-10pt}
\end{figure}

We define a curriculum using the generalized logistic function~\citep{richards1959flexible} of the form:
\begin{equation}\label{eq:glf}
    w(t; r, s) = \frac{1}{1+\exp(-r \times (t-s))},
\end{equation}
where $r\in\R$ is the rate-of-change parameter, which specifies how fast the weight can increase ($r>0$) or decrease ($r<0$); $t\in[0,1]$ is the training progress (typically iteration number divided by max iterations); and $s\in\R$ shifts the pivot weight of the logistic function ($w(.)=.5$) to the left or right such that at $t=s$ the weight is $0.5$. Figure~\ref{fig:base_sigmoid} illustrates the effect of these parameters. Greater absolute values for the rate parameter enforce faster rates of change in weights, while greater values of the shift parameter enforce longer delays in reaching the pivot weight of $0.5$. These parameters provide flexibility in controlling sample weights during training, which is key for deriving effective curricula. 
%
The above function can approximate existing predefined curricula. For example, Figure~\ref{fig:inc_cfg} shows a specific configuration for the logistic functions for standard CL~\citep{Bengio2009}, where training starts with easier samples and gradually proceeds with harder ones.

\subsection{Non-monotonic Curricula}\label{sec:dyn_class}
Although the generalized logistic function in (\ref{eq:glf}) can lead to effective curricula, {\em monotonic} functions are limited in their coverage capacity. For example, they do not allow easy samples with low weights to become important again (receive high weights) at later stages of training to mitigate {\em forgetting}, which is a major challenge for effective curriculum learning~\citep{forgetting2019,zhou2020curriculum}. 

We address this challenge by extending the framework to non-monotonic curricula, where samples can \emph{move} between difficulty classes based on their {\em learning progress} during training. We quantify learning progress for training samples based on the deviation of their losses from the average losses of their corresponding difficulty groups.
At every iteration, samples with loss values greater 
than the average are {\em promoted} to their immediate higher difficulty groups and 
the rest are {\em demoted} to their immediate lower difficulty groups. 
These movements allow monotonic weight functions result in non-monotonic and multimodal weight trajectories for training samples, which improves the search capability of our framework and addresses the forgetting challenge.

\begin{figure*}[htb]
    \centering
     \begin{subfigure}{0.20\linewidth}
         \centering
         \includegraphics[width=\linewidth]{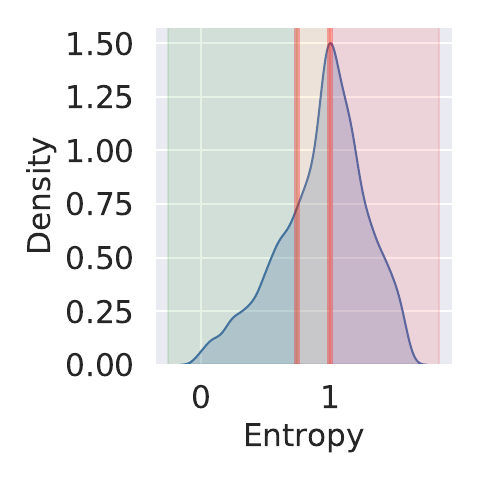}
         \caption{ChaosNLI Entropy}
     \end{subfigure}
     \begin{subfigure}{0.20\linewidth}
         \centering
         \includegraphics[width=\linewidth]{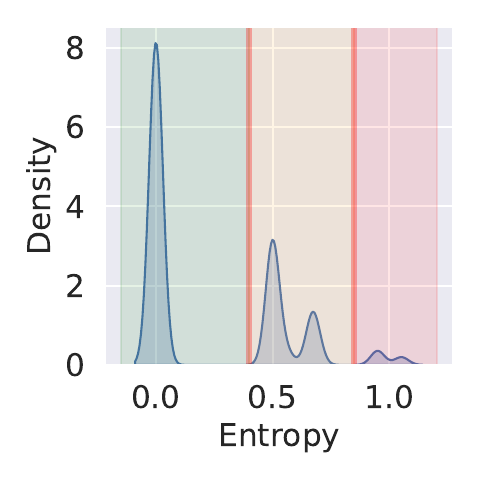}
         \caption{SNLI Entropy}
         \label{fig:snli_ent}
     \end{subfigure}
     \begin{subfigure}{0.20\linewidth}
         \centering
         \includegraphics[width=\linewidth]{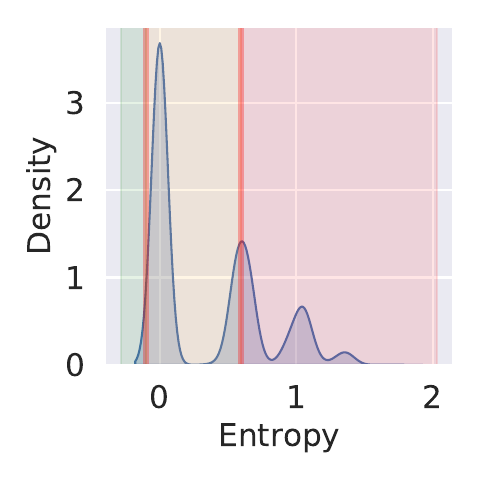}
         \caption{Twitter Entropy}
     \end{subfigure}
     \begin{subfigure}{0.20\linewidth}
         \centering
         \includegraphics[width=\linewidth]{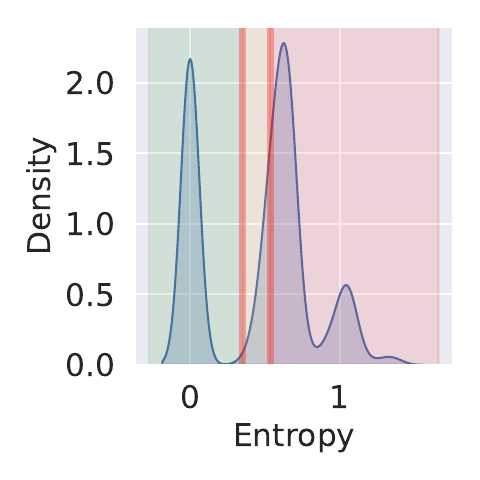}
         \caption{Reddit Entropy}
     \end{subfigure}
     
     \begin{subfigure}{0.20\linewidth}
         \centering
         \includegraphics[width=\linewidth]{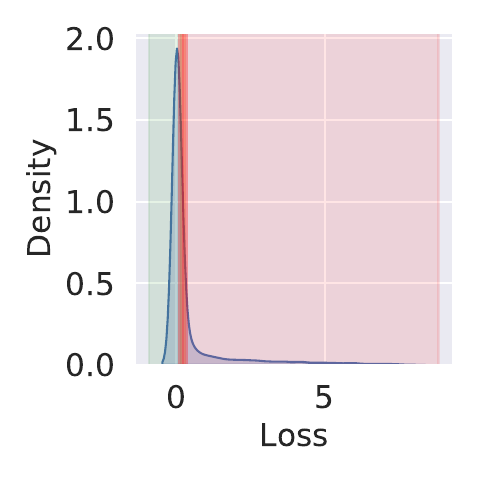}
         \caption{ChaosNLI Loss}
     \end{subfigure}
     \begin{subfigure}{0.20\linewidth}
         \centering
         \includegraphics[width=\linewidth]{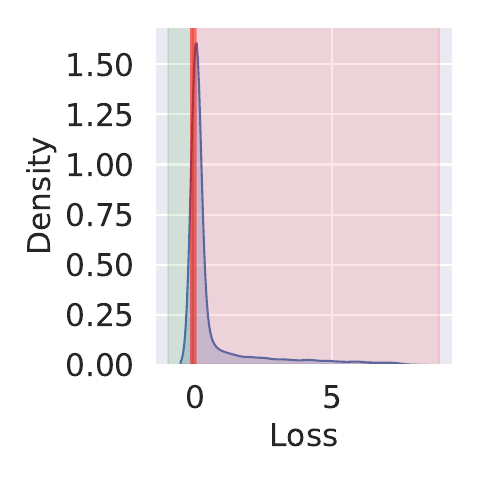}
         \caption{SNLI Loss}
        \label{fig:snli_loss}
     \end{subfigure}
     \begin{subfigure}{0.20\linewidth}
         \centering
         \includegraphics[width=\linewidth]{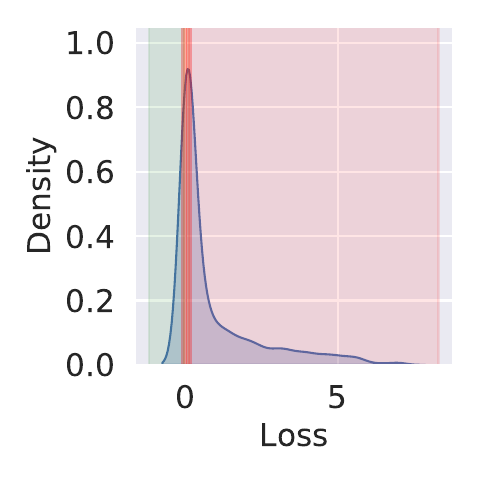}
         \caption{Twitter Loss}
     \end{subfigure}
     \begin{subfigure}{0.20\linewidth}
         \centering
         \includegraphics[width=\linewidth]{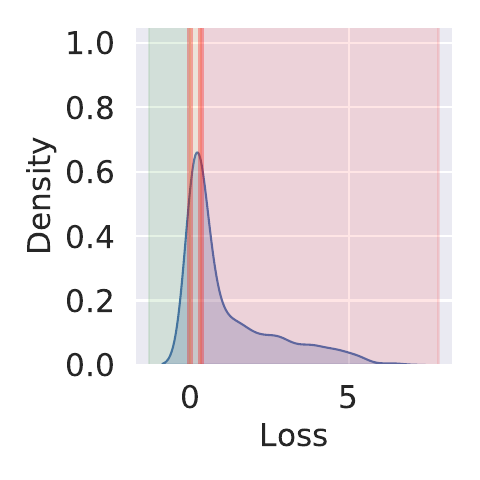}
         \caption{Reddit Loss}
     \end{subfigure}
    \caption{Distributions of entropy and loss in our datasets. Samples of the {\em easy} class are to the left of the first vertical line and shaded in green, those of the {\em medium} class are between the two vertical lines and shaded in orange, and samples of the {\em hard} class are to the right of the second line and shaded in red.}
    \label{fig:diff_classes}
\end{figure*}

\subsection{Parameter Optimization}\label{sec:discovery}
We find the optimal
curriculum parameters $(r,s)$ for each difficulty group using the 
Tree-structured Parzen Estimator (TPE) algorithm~\citep{bergstra2011algorithms,  akiba2019optuna}, which, unlike the grid or random search, traverses the parameter space by estimating the parameters that are most probable to perform better on a trial. 
Using this method, we can learn data-driven curricula beyond what could be manually designed through empirical settings or choices among the limited ordering strategies.

The discovered curricula are optimal within our search space, as defined by the weight functions and searchable parameters. However, in practice, we observed that the change in performance across the missing regions in the search space
is minor. Given that our weight functions can approximate other curricula learned by existing CL models, see \S\ref{sec:encompass}, we expect the optimum curriculum within our search space closely approximates the optimal curriculum for each dataset and model pair.

\subsection{Prior Knowledge of Difficulty} \label{sec:scoring}
Annotation entropy is a natural measure of difficulty (for humans) and may serve as a reliable difficulty metric for models. Entropy of each sample $x_i$ is calculated as $-\sum_l p_c \log{p_c}$~\citep{shannon1948mathematical}, where $c$ is a class category and $p_c$ is the fraction of annotators who chose label $c$ for the sample. The use of entropy is supported in~\citep{nie2020can}, reporting a consistent positive correlation between model accuracy and level of human agreement.

Furthermore, moving average of a sample's instantaneous loss is a good metric for difficulty \cite{zhou2020curriculum}. Using a baseline model trained with no curriculum and with default hyper-parameters, we collect the loss values of all training instances at intervals of 0.5 epochs and use the average loss as prior knowledge about sample difficulty. We obtain twenty observations of the loss and compute the average for each sample. 

Figure~\ref{fig:diff_classes} shows the distributions of entropy and loss, and examples of data partitions across four datasets. Most datasets are highly imbalanced across difficulty groups, often containing more easier samples than harder ones. Such data disparities would perhaps explain why 
computational models can achieve human-level performance on complex NLP tasks or recent results reporting neural models being largely invariant to random word order permutation of data~\citep{sinha-etal-2021-unnatural}. 

We acknowledge that while multiple annotations per sample may not be readily available for many NLP datasets, such annotations were collected for most NLP datasets at their dataset development time. Our work shows that such information can be used to find effective curricula for NLP models and encourages dataset creators to publish their full annotation information. In addition, our curriculum discovery framework is independent of annotation information. In fact, we evaluated our approach with both annotation entropy and loss as two choices for sample-level difficulty estimation.

\section{Experiments}
\subsection{Datasets} \label{sec:datasets}
For the purpose of our experiments, we chose datasets for which several annotations per sample are available. Such annotator-level information is often available at the creation time of most NLP datasets and provide rich information for effective learning. Before training, we partition each dataset into $k$ difficulty groups using $\{\frac{i}{k}\}_{i=0}^{i=k}$ quantiles.

{\bf SNLI}~\citep{bowman2015large}. The Stanford Natural Language Inference (SNLI) benchmark~\citep{bowman2015large} contains 
36.7k and 2.6k samples annotated by 5 and 4 workers respectively, which we refer to as SNLI full in our experiments.  

{\bf ChaosNLI}~\citep{nie2020} contains 100 annotations per sample for about 1.5K development samples of SNLI and
MNLI~\citep{williams2017broad}. We use these samples as training data, the remaining 8.5K development samples of SNLI as development set, and the test set of SNLI as test set. 

{\bf Twitter}~\citep{amiri2018toward}. This dataset has been developed to obtain population-level statistics of alcohol use reports through social media.
It contains more than 9k 
tweet, annotated by at least three workers for report of first-person alcohol use, intensity of the drinking (light vs. heavy), context of drinking (social vs. individual), and time of drinking (past, present, or future). We define a multi-class classification task for this dataset based on the above categories, see the data distribution in Appendix~\ref{sec:data_freq}. We randomly split the data into 5.4k, 1.8k and 1.8k training, development and test sets. 

{\bf Reddit.} We developed this dataset to obtain population-level statistics of cancer patients. It contains 3.8k Reddit posts annotated by at least three annotators for relevance to specific cancer types. 
We define a multi-class classification task based on post relevance and cancer type, see 
Appendix~\ref{sec:data_freq}. We randomly split the data into 2.2k, 765, and 765 training, development and test sets respectively. 

ChaosNLI is balanced in its difficulty groups. We create {\em difficulty-balanced} versions of SNLI, Twitter and Reddit by collecting an equal number of samples from each difficulty group. The resulting datasets contain 1.7K to 2.3K samples.



\subsection{Baselines}\label{sec:baselines}

\paragraph{No-CL} The conventional training approach, which involves utilizing all samples for training in each iteration.

\paragraph{Self-paced Learning (SPL)}~\cite{kumar2010self} weights instances based on their difficulty to the model by optimizing the following objective: 
\begin{equation}
    \mathcal{L}(\mathcal{D}; \theta) =\arg\min_{\bm{v}} \sum_i^n v_i l_i + f(\bm{v}; \lambda),
\end{equation}
where $l_i$ is the loss of instance $i$ parameterized by $\theta$, $v_i$ is a trainable weight parameter assigned to each instance, and $f$ is a regularization function for the weights. 
The model finds $\textbf{v}$ that minimizes its loss under the constraint of $f$. 
The binary scheme SPL is defined by the regularization function $f(\textbf{v}; \lambda) = - \lambda \|\textbf{v}\|_1$;
if $l_i < \lambda$, $v_i = 1$, otherwise $v_i = 0$, i.e., only easy samples are selected at each step. 

\paragraph{Mentornet}~\cite{jiang2018mentornet} uses an auxiliary network to weight samples at every iteration. The network takes as input recent loss history, 
running mean of the loss, current epoch number (to account for training progress), and target labels. The network consists of an LSTM layer to encode the $k$ steps of loss, embedding matrices for the target label and epoch number; a fully connected layer; and a final sigmoid layer. The sigmoid layer outputs weights of samples for training.

\paragraph{Difficulty Prediction (DP)}~\cite{yang2019predicting} defines sample difficulty 
as follows: 
\begin{equation}
    d_i = \frac{\sum_{j = 1}^{l_i}f(y_i^{(j)}, \hat{y}_i)}{l_i},
    \label{eq:dp_d}
\end{equation}
where $\hat{y}_i$ is the ground truth label and $f$ measures the Spearman's rank correlation coefficient between labels produced by experts and non-experts. The model re-weights samples for performance improvement using a pre-defined threshold $\tau$,:
\begin{equation}
    1 - \alpha \frac{d_i - \tau}{1 - \tau}.
    \label{eq:dp_weight}
\end{equation}

\paragraph{SuperLoss (SL)}~\cite{castells2020superloss} uses the following function 
to estimate sample weights:
\begin{equation}
    \mathcal{L}_\lambda = (l_i - \tau) \: \sigma_i + \lambda \: (\log{\sigma_i})^2,
    \label{eq:sl_obj}
\end{equation}
where $\tau$ is the moving average of loss (as the measure of difficulty) and $\sigma$ is sample confidence.
The model emphasizes easy samples (those with small losses) 
throughout the training.

Our approach employs two difficulty scoring functions and two curriculum types for each dataset. 
The difficulty scoring functions are {\em Loss} and {\em Ent} (entropy) described in \S\ref{sec:scoring}.
The first curriculum type ({\em inc}) is the off-the-shelf gradually increasing approach in Figure~\ref{fig:inc_cfg}, which is rapidly computed and applied to all models, resulting in \textbf{Ent(inc)} and \textbf{Loss(inc)} approaches. The non-monotonic version of the {\em inc} curriculum (\S\ref{sec:dyn_class}) are labeled \textbf{Ent+(inc)} and \textbf{Loss+(inc)}. The second curriculum type ({\em sp}, for specialized) is obtained through the proposed optimization approach (\S\ref{sec:discovery}) that finds optimal curricula for each model and dataset, resulting in \textbf{Ent(sp)} and \textbf{Loss(sp)}. 


 \begin{table*}[t]
    \centering
    \begin{tabular}{l ccc cccc c}
        & \multicolumn{3}{c}{{\bf Full}} & \multicolumn{4}{c}{{\bf Difficulty Balanced}} \\
        \toprule
        & \textbf{SNLI} & \textbf{Twitter} & \textbf{Reddit} & \textbf{ChaosNLI} & \textbf{SNLI} & \textbf{Twitter} & \textbf{Reddit} & \textbf{Avg}\\
        \toprule
        
\textbf{Ent (sp)} & {\small \textbf{88.3}} {\tiny $\pm$ 0.04} & {\small 79.1} {\tiny $\pm$ 0.15} & {\small 73.5} {\tiny $\pm$ 0.22} & {\small \textbf{78.3}} {\tiny $\pm$ 0.49} & {\small 80.6} {\tiny $\pm$ 0.16} & {\small 76.7} {\tiny $\pm$ 0.14} & {\small 72.4} {\tiny $\pm$ 0.46} & {\small \textbf{78.4}}\\

\textbf{Ent (inc)} & {\small 88.0} {\tiny $\pm$ 0.05} & {\small 79.4} {\tiny $\pm$ 0.11} & {\small 73.5} {\tiny $\pm$ 0.21} & {\small 77.5} {\tiny $\pm$ 0.64} & {\small 80.6} {\tiny $\pm$ 0.25} & {\small 76.7} {\tiny $\pm$ 0.17} & {\small 71.1} {\tiny $\pm$ 0.22} & {\small 78.0}\\

\textbf{Ent+ (inc)} & {\small 88.0} {\tiny $\pm$ 0.17} & {\small \textbf{79.7}} {\tiny $\pm$ 0.17} & {\small \textbf{73.9}} {\tiny $\pm$ 0.21} & {\small 77.8} {\tiny $\pm$ 0.39} & {\small 77.9} {\tiny $\pm$ 2.10} & {\small \textbf{77.2}} {\tiny $\pm$ 0.18} & {\small 72.9} {\tiny $\pm$ 0.28} & {\small 78.2} \\

\textbf{Loss (sp)} & {\small 88.0} {\tiny $\pm$ 0.05} & {\small 79.3} {\tiny $\pm$ 0.17} & {\small 72.6} {\tiny $\pm$ 0.23} & {\small 76.8} {\tiny $\pm$ 0.90} & {\small \textbf{81.4}} {\tiny $\pm$ 0.16} & {\small 77.0} {\tiny $\pm$ 0.16} & {\small 73.0} {\tiny $\pm$ 0.61} & {\small 78.3}\\

\textbf{Loss (inc)} & {\small 87.9} {\tiny $\pm$ 0.06} & {\small 78.9} {\tiny $\pm$ 0.11} & {\small 72.7} {\tiny $\pm$ 0.16} & {\small 74.7} {\tiny $\pm$ 0.86} & {\small 80.8} {\tiny $\pm$ 0.37} & {\small 75.7} {\tiny $\pm$ 0.19} & {\small 71.7} {\tiny $\pm$ 0.69} & {\small 77.5}\\

\textbf{Loss+ (inc)} & {\small 87.8} {\tiny $\pm$ 0.09} & {\small 78.6} {\tiny $\pm$ 0.31} & {\small 72.3} {\tiny $\pm$ 0.48} & {\small 74.0} {\tiny $\pm$ 1.26} & {\small 79.0} {\tiny $\pm$ 0.91} & {\small 76.6} {\tiny $\pm$ 0.36} & {\small \textbf{73.0}} {\tiny $\pm$ 0.34} & {\small 77.3}\\

\midrule
\textbf{DP} & {\small 88.1} {\tiny $\pm$ 0.06} & {\small 78.5} {\tiny $\pm$ 0.12} & {\small 73.0} {\tiny $\pm$ 0.24} & {\small 76.4} {\tiny $\pm$ 0.22} & {\small 79.6} {\tiny $\pm$ 0.36} & {\small 76.1} {\tiny $\pm$ 0.15} & {\small 71.5} {\tiny $\pm$ 0.35} & {\small 77.6}\\

\textbf{SL} & {\small 88.0} {\tiny $\pm$ 0.07} & {\small 78.6} {\tiny $\pm$ 0.13} & {\small 73.1} {\tiny $\pm$ 0.24} & {\small 77.3} {\tiny $\pm$ 0.53} & {\small 78.2} {\tiny $\pm$ 0.48} & {\small 76.0} {\tiny $\pm$ 0.15} & {\small 70.7} {\tiny $\pm$ 0.41} & {\small 77.4}\\

\textbf{MentorNet} & {\small 87.7} {\tiny $\pm$ 0.18} & {\small 78.2} {\tiny $\pm$ 0.12} & {\small 73.1} {\tiny $\pm$ 0.23} & {\small 76.0} {\tiny $\pm$ 0.00} & {\small 79.0} {\tiny $\pm$ 0.69} & {\small 76.3} {\tiny $\pm$ 0.16} & {\small 71.1} {\tiny $\pm$ 0.48} & {\small 77.3}\\

\midrule
\textbf{No-CL} & {\small 87.9} {\tiny $\pm$ 0.07} & {\small 78.6} {\tiny $\pm$ 0.12} & {\small 73.3} {\tiny $\pm$ 0.20} & {\small 76.2} {\tiny $\pm$ 0.27} & {\small 79.4} {\tiny $\pm$ 0.32} & {\small 76.4} {\tiny $\pm$ 0.16} & {\small 70.8} {\tiny $\pm$ 0.26} & {\small 77.5}\\
\bottomrule

\end{tabular}
    \caption{{\em Loss} and {\em Ent} indicate curricula that partition the data based on $k=3$ difficulty groups determined by loss and entropy respectively, see \S\ref{sec:scoring}. {\em inc} is the easy to hard curriculum shown in Figure \ref{fig:inc_cfg}. {\em sp} is the specialized curriculum obtained by curriculum discovery, see \S\ref{sec:discovery}, which is different for each dataset.}
    \label{tab:acc}
\end{table*}

\subsection{Settings}
\label{sec:setting}
We use bayesian optimization to tune the parameters $\lambda$ of SL and $\alpha$ and $\tau$ of DP on development data. The optimal values found are $\lambda = 1.2$, $\alpha = 0.9$ and $\tau$ is set dynamically upon loading the dataset to the 50 percentile difficulty value of the training data. 
%
We use \textit{twitter-roberta-base} for Twitter and \textit{roberta-base} for other datasets, both from~\citep{wolf-etal-2020-transformers}. 
We set learning rate to $1 \times 10^{-5}$, batch size to $16$, epochs to $10$ (we confirm that this number of iterations is sufficient for all models to converge), and use Adam optimizer~\citep{kingma2017adam}. The checkpoint with the best performance is used for testing. For each experiment, we train the model 
using five random seeds 
and report standard error.

In addition, 
we set the search space for the rate ($r$) and shift ($s$) parameters to $[-10, 10]$ with a step of $2$ and $[-0.5, 1.5]$ with a step of $0.25$ respectively. 
The search is run for at least 100 trials using the method described in (\S\ref{sec:discovery}). Each trial is run with three seeds and the result is averaged. The search objective is to maximize accuracy over development data. The trial number in which the best parameters are found is reported in Appendix~\ref{sec:computation}.
We only search for curricula with three difficulty groups to ease interpretability and improve readability, and to minimize the number of search parameters. However, in case of {\em inc} curriculum, the optimal number of difficulty groups for ChaosNLI, SNLI, Twitter, Reddit are 12, 3, 28, and 12 respectively; in all cases, we tune the number of groups on the development set and evaluate on the best performing one. Appendix~\ref{sec:fg} includes the results of tuning the number of groups.

\subsection{Curriculum Discovery Improves Models}
Table~\ref{tab:acc} shows that the gradually increasing curriculum using entropy, {\em Ent (inc)}, achieves better accuracy than {\em No-CL} and other baselines, and the difference is significant. The gain is often greater with more than 3 difficulty groups, see detail results in Figure~\ref{fig:fg}, Appendix~\ref{sec:fg}. Both ({\em inc}) and the specialized ({\em sp}) curricula often perform better than the baselines. On average, entropy as scoring function performs better than loss, indicating prior knowledge based on difficulty to humans is useful to the model. The results also show that non-monotonic curricula (Ent+, Loss+) can further improve the performance; we attribute this result to the ability of the non-monotonic curricula to dynamically adjust the difficulty of samples according to model behavior as training progresses, allowing easier or harder samples to the model accumulate in the easier and harder difficulty groups. The performance improvement is more pronounced on the difficulty balanced datasets compared to full datasets, which can be attributed to the balanced nature or smaller size of these datasets.

\begin{figure}[!ht]
     \centering
     
     \begin{subfigure}[htb]{0.23\linewidth}
         \centering
         \includegraphics[width=\linewidth]{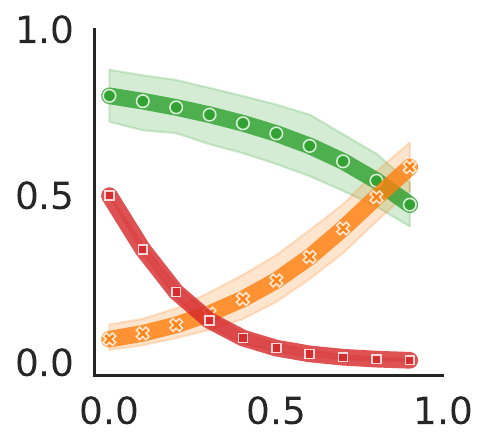}
         \caption{S-D-E}
     \end{subfigure}
     \begin{subfigure}[htb]{0.23\linewidth}
         \centering
         \includegraphics[width=\linewidth]{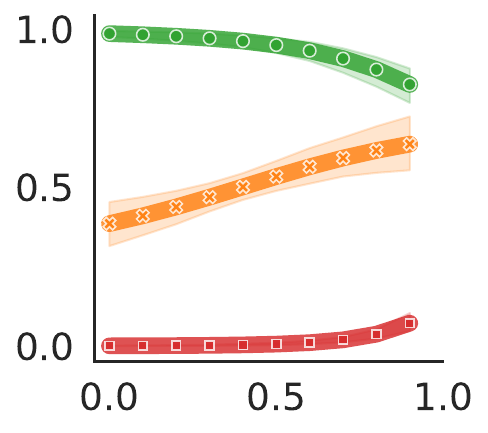}
         \caption{S-D-L}
         \label{fig:sdl_cfg}
     \end{subfigure}
     \begin{subfigure}[htb]{0.23\linewidth}
         \centering
         \includegraphics[width=\linewidth]{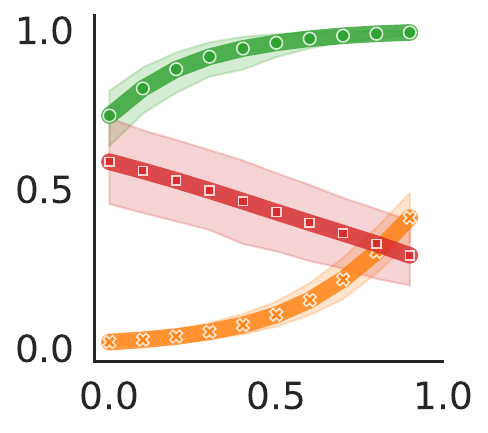}
         \caption{S-F-E}
         \label{fig:sfe_cfg}
     \end{subfigure}
     \begin{subfigure}[htb]{0.23\linewidth}
         \centering
         \includegraphics[width=\linewidth]{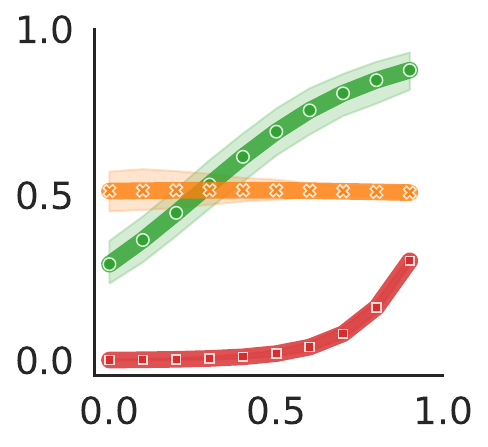}
         \caption{S-F-L}
     \end{subfigure}
     
     \begin{subfigure}[htb]{0.23\linewidth}
         \centering
         \includegraphics[width=\linewidth]{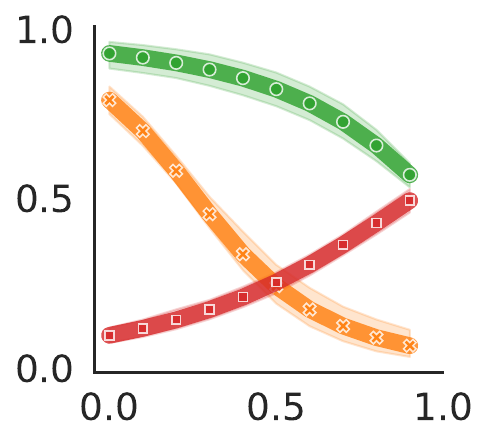}
         \caption{T-D-E}
     \end{subfigure}
     \begin{subfigure}[htb]{0.23\linewidth}
         \centering
         \includegraphics[width=\linewidth]{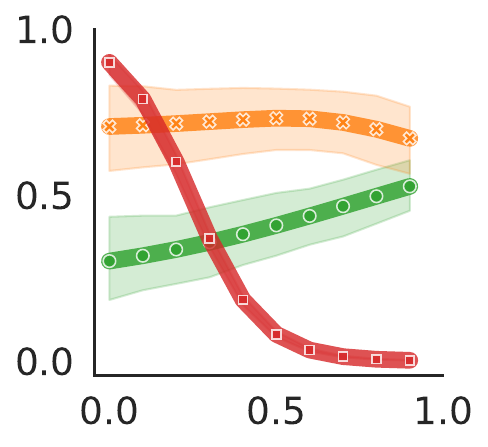}
         \caption{T-D-L}
         \label{fig:tdl_cfg}
     \end{subfigure}
     \begin{subfigure}[htb]{0.23\linewidth}
         \centering
         \includegraphics[width=\linewidth]{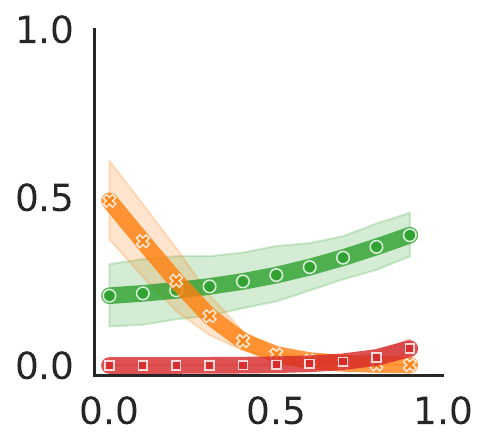}
         \caption{T-F-E}
         \label{fig:afe_cfg}
     \end{subfigure}
     \begin{subfigure}[htb]{0.23\linewidth}
         \centering
         \includegraphics[width=\linewidth]{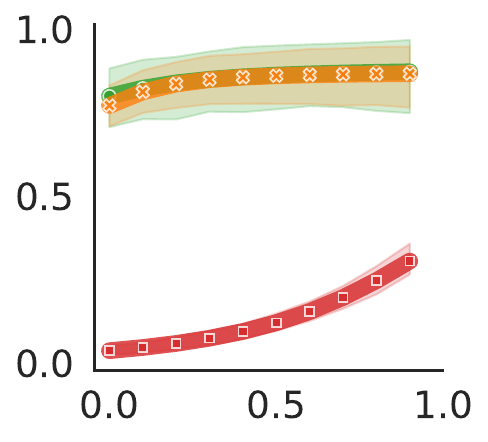}
         \caption{T-F-L}
     \end{subfigure}
     
     \begin{subfigure}[htb]{0.23\linewidth}
         \centering
         \includegraphics[width=\linewidth]{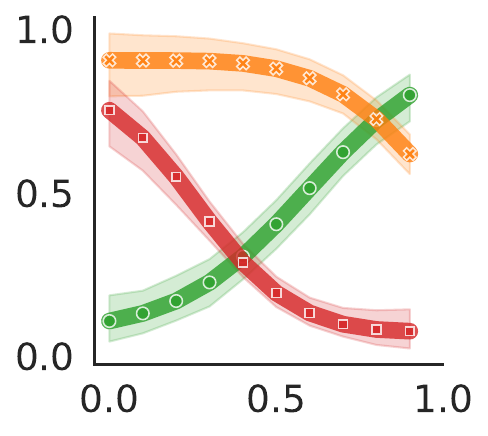}
         \caption{R-D-E}
     \end{subfigure}
     \begin{subfigure}[htb]{0.23\linewidth}
         \centering
         \includegraphics[width=\linewidth]{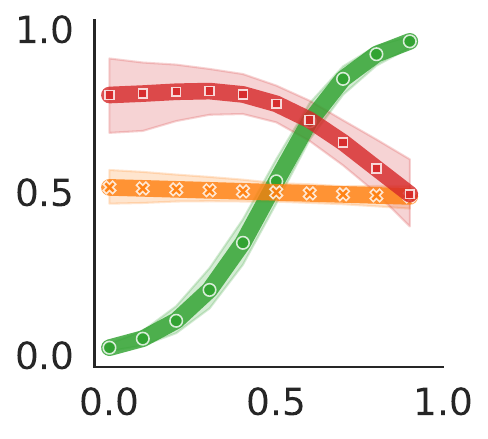}
         \caption{R-D-L}
     \end{subfigure}
     \begin{subfigure}[htb]{0.23\linewidth}
         \centering
         \includegraphics[width=\linewidth]{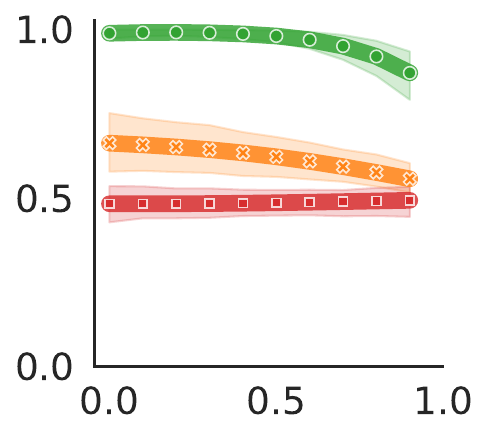}
         \caption{R-F-E}
     \end{subfigure}
     \begin{subfigure}[htb]{0.23\linewidth}
         \centering
         \includegraphics[width=\linewidth]{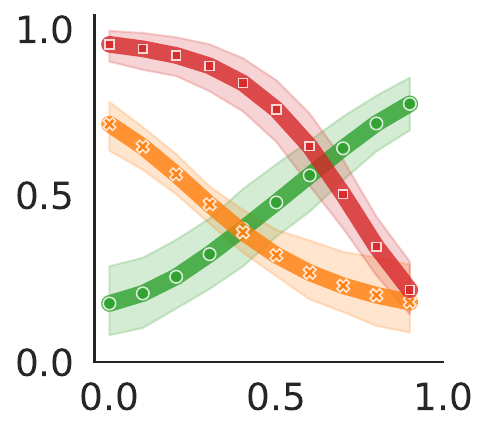}
         \caption{R-F-L}
     \end{subfigure}
     
     \begin{subfigure}[htb]{0.23\linewidth}
         \centering
         \includegraphics[width=\linewidth]{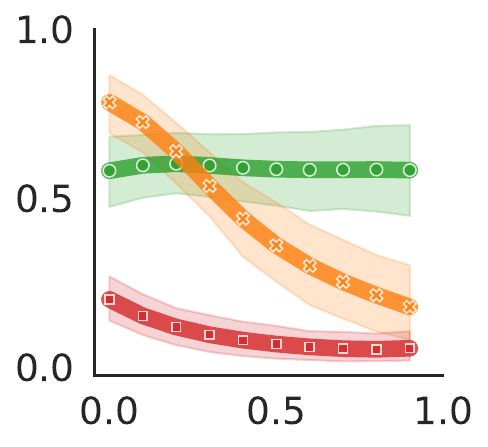}
         \caption{C-D-E}
     \end{subfigure}
     \begin{subfigure}[htb]{0.23\linewidth}
         \centering
         \includegraphics[width=\linewidth]{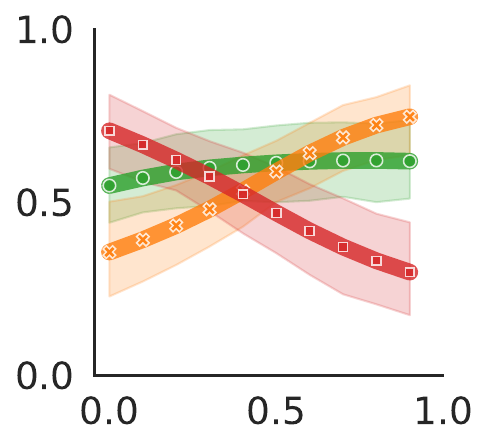}
         \caption{C-D-L}
     \end{subfigure}
     \hspace{0.48\linewidth}
     \caption{Each caption is composed of the first character of the name of a dataset: \{\textbf{C}haosNLI, \textbf{S}NLI, \textbf{T}witter, \textbf{R}eddit\}, followed by the type of the dataset \{\textbf{D}ifficulty-balanced or \textbf{F}ull\}, and the difficulty score used \{\textbf{E}ntropy, \textbf{L}oss\} in experiments. The x-axis is the training progress and y-axis is the confidence assigned to samples of a difficulty-class. The green line (circle marker) is {\em easy}, orange line (x marker) is {\em medium}, and red line (diamond marker) is {\em hard}. The solid line is the mean of the top 25 performing configurations for each dataset and scoring function pair, and the shaded area represents the 95\% CI.}
     \vspace{-10pt}
    \label{fig:configs}
\end{figure}

\begin{figure*}[!ht]
    \centering
    \includegraphics[width=0.8\linewidth]{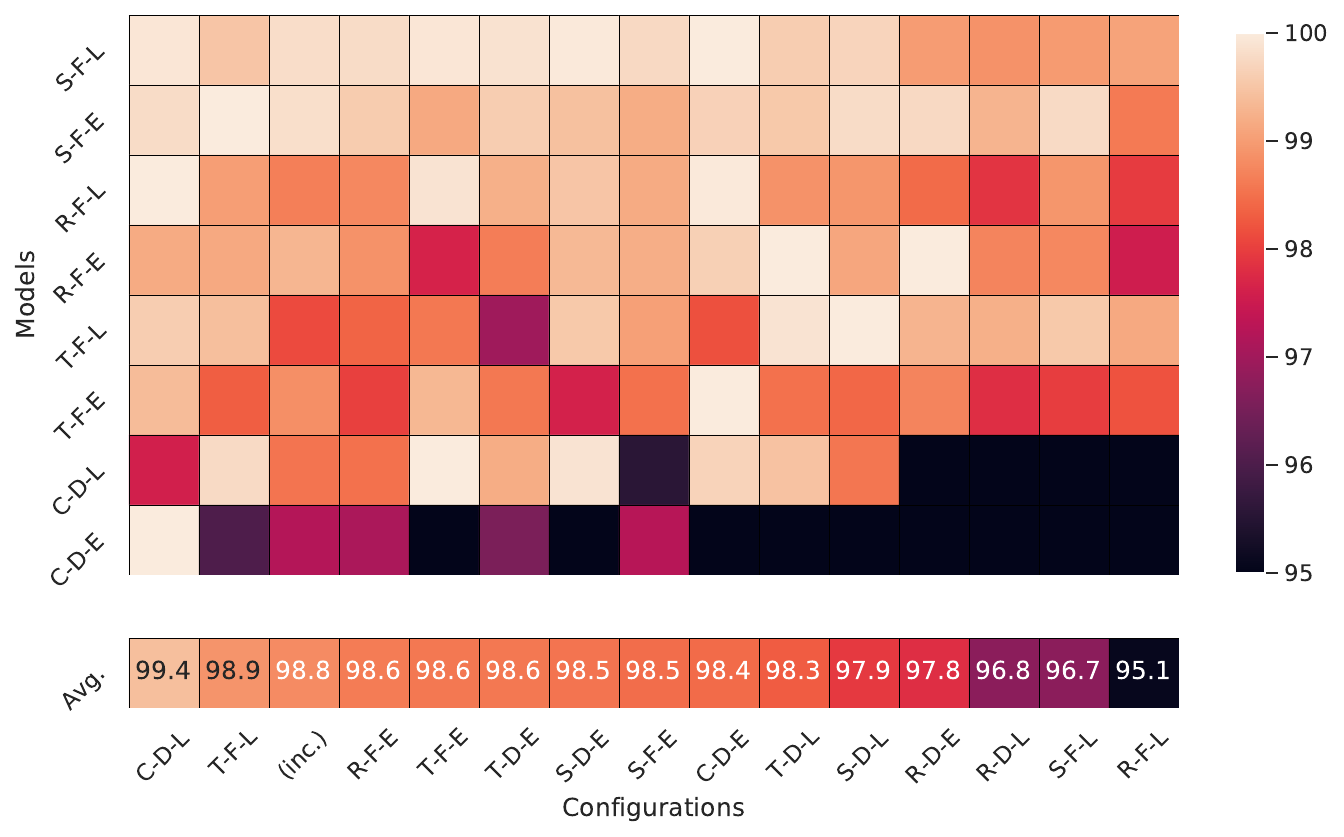}
    \caption{Notation is the same as Figure~\ref{fig:configs}: \{\textbf{C}haosNLI, \textbf{S}NLI, \textbf{T}witter, \textbf{R}eddit\}, followed by the type of the dataset \{\textbf{D}ifficulty-balanced or \textbf{F}ull\}, and the difficulty score used \{\textbf{E}ntropy, \textbf{L}oss\}. The x-axis lists curricula discovered using a particular dataset and scoring function, and the increasing curriculum {\em inc} (Figure~\ref{fig:inc_cfg}). The y-axis lists models that are trained using each curriculum. For example, the cell at the intersection of row "S-F-L" and column "T-F-E" represents a model trained on SNLI full partitioned by loss, using the curriculum discovered for the full Twitter dataset partitioned by entropy (Figure~\ref{fig:afe_cfg}). Each row of the Table is normalized to match the scales of different models (after normalization, the max of each row is 100).}
    \label{fig:nxn}
    \vspace{-15pt}
\end{figure*}

\subsection{Discovered Curricula Are Non-monotonic}
Figure~\ref{fig:configs} shows the mean and 95\% CI of the top 25 performing curricula.
The resulting curricula are non-monotonic and greatly differ from the known strategies reported in literature, such as gradually increasing difficulty or anti-curriculum. In addition, the weights of hard samples tend to decrease, supporting the hypothesis that these instances may be too difficult or noisy for models to learn. 
In addition, in SNLI and Twitter {\em easy} samples often carry the most significant weight, unlike Reddit, where {\em easy} samples are often down-weighted early during the training. These weighting patterns reveal the relative importance of samples in each dataset.
Finally, the full SNLI dataset with entropy partitions provides useful information. In Figure~\ref{fig:sfe_cfg}, {\em hard} samples are assigned weights around 0.5, unlike the three other cases of SNLI. We attribute this result to the reduced presence of {\em hard} samples (skewed entropy in Figure~\ref{fig:snli_ent}).


\subsection{Discovered Curricula Are Generalizable}
\label{sec:generalizable}

Figure~\ref{fig:nxn} shows the accuracy obtained when the top-performing discovered curriculum for one dataset (from Figure~\ref{fig:configs}) is applied to other datasets. Each cell is the average result of 5 seeds. We observe common characteristics among datasets that cause the curriculum to be transferable between them.
First, the top generalizable configuration is obtained from ChaosNLI, the dataset with the richest inter-annotator entropy signal. Therefore, the quality of the difficulty score is important to the discovery of an effective curriculum.
Second, the {\em inc} configuration is among the most generalizable configurations, with no added cost in its creation.
Third, the curricula obtained using the small, down-sampled difficulty-balanced datasets generalize well and achieve high performance on the large datasets. This is useful as 
curriculum discovery is much faster on smaller datasets, and the framework can be applied to large datasets by searching for a curriculum on a small subset of the data, mitigating the computational expenses of 
using full datasets.
Fourth, as noted previously, instances of the Reddit dataset consist of long paragraphs, causing high variance in models trained using the dataset. Consequently, the curricula obtained using the Reddit and loss as measure of difficulty are of lower quality and perform poorly. Appendix~\ref{sec:nxn_full} reports the results of all configurations. 

Table~\ref{tab:gen-size} shows the transferability of discovered curricula across model sizes. We consider three models with increasing sizes applied to ChaosNLI: {\tt distilroberta-base} with 82M parameters, {\tt roberta-base} with 125M parameters, and {\tt bart-large} with 406M parameters. The results show that the curricula discovered for small models 
are transferable to larger models, with significant improvement over No-CL and other CL baselines. In particular, we observe greater transferability for smaller model sizes, which indicates curriculum discovery is more beneficial to smaller models than larger (more robust) models. In some cases, the curricula discovered for smaller models perform better than those discovered for larger models, see Ent(sp)~{\small 82M}~and~{\small 125M}. This is because curriculum discovery is less expensive on smaller models, allowing better exploration of curriculum space to find better curricula. 

\begin{figure}[htb]
    \centering
     \begin{subfigure}{0.63\linewidth}
         \centering
         \includegraphics[width=\linewidth]{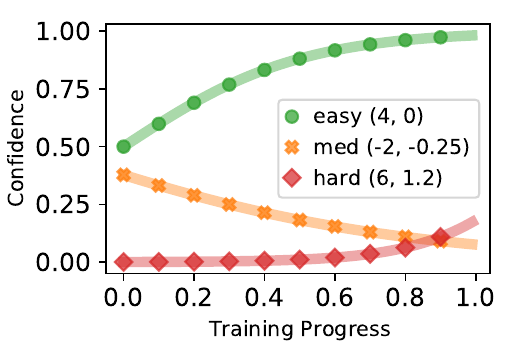}
         \caption{82M Parameter Model}
     \end{subfigure}
     \quad
     \begin{subfigure}{0.63\linewidth}
         \centering
         \includegraphics[width=\linewidth]{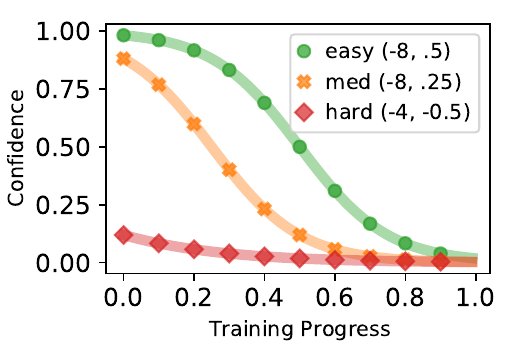}
         \caption{125M Parameter Model}
     \end{subfigure}
     \quad
     \begin{subfigure}{0.63\linewidth}
         \centering
         \includegraphics[width=\linewidth]{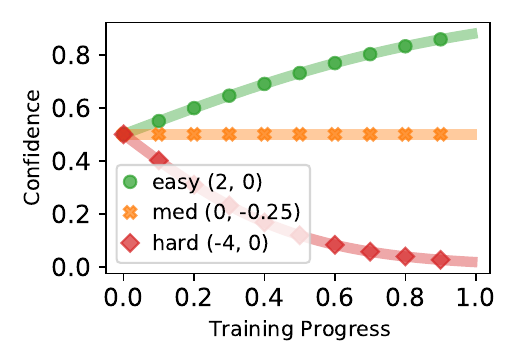}
         \caption{406M Parameter Model}
     \end{subfigure}
    \caption{Specialized curricula optimized on ChaosNLi using {\tt distilroberta} (82M), {\tt roberta-base} (125M), and {\tt facebook/bart-large} (406M). The performances of each curriculum are reported in Table~\ref{tab:gen-size}.}
    \label{fig:size-cfgs}
\end{figure}

Figure~\ref{fig:size-cfgs} shows the curricula obtained using models of different sizes. The three curricula are similar in their relative treatment of difficulty groups: samples from the easy class are assigned higher weights than those from the medium class, and medium samples receive higher weights than hard samples. In addition, hard samples are considerably down-weighted, which indicates deemphasizing hard samples during training can lead to better results on the test data of ChaosNLi.


\begin{table}[t]
    \centering
    \begin{tabular}{l|l|l|l}
    {\bf Curriculum} & {\bf 82M} & {\bf 125M} & {\bf 406M} \\
    \toprule
    No-CL & {\small 63.9} {\tiny $\pm$ 0.13} & {\small 76.2} {\tiny $\pm$ 0.27} & {\small 80.0} {\tiny $\pm$ 0.41} \\
    Best baseline & {\small 64.7} {\tiny $\pm$ 0.3} & {\small 77.3} {\tiny $\pm$ 0.53} & {\small 81.9} {\tiny $\pm$ 0.86} \\
    \midrule
    Ent (sp) {\small 82M} & \textbf{{\small 67.4}} {\tiny $\pm$ 0.25} & \textbf{{\small 78.4}} {\tiny $\pm$ 0.46} & {\small 81.5} {\tiny $\pm$ 0.50} \\
    Ent (sp) {\small 125M} & -- & {\small 78.3} {\tiny $\pm$ 0.49} & \textbf{{\small 82.6}} {\tiny $\pm$ 0.39}\\
    Ent (sp) {\small 406M} & -- & -- & {\small 82.3} {\tiny $\pm$ 0.54} \\
    \end{tabular}
    \caption{Transferability of the specialized curricula discovered for small models to large models on ChaosNLI. ``Best baseline'' shows the best performance obtained by baselines (DP, SL, Mentornet). ``Ent (sp) {\small $n$}'' indicates the curriculum discovered on the model with $n$ parameters. Column headers indicate the model trained using  the discovered curricula of the corresponding rows.}
    \label{tab:gen-size}
    \vspace{-10pt}
 \end{table}

\subsection{Potential to Encompass Existing Models}
\label{sec:encompass}
The framework presented in this paper is capable of representing curriculum learning approaches that prune noisy data, e.g.~\citep{northcutt2021confident}, 
use different sub-samples of data during training, e.g.~\citep{xu2020curriculum}, and
re-weight loss according to sample difficulty, choosing to emphasize either easy or hard samples, e.g.~\citep{castells2020superloss}.  

\begin{figure}[htb]
    \centering
     \begin{subfigure}{0.47\linewidth}
         \centering
         \includegraphics[width=\linewidth]{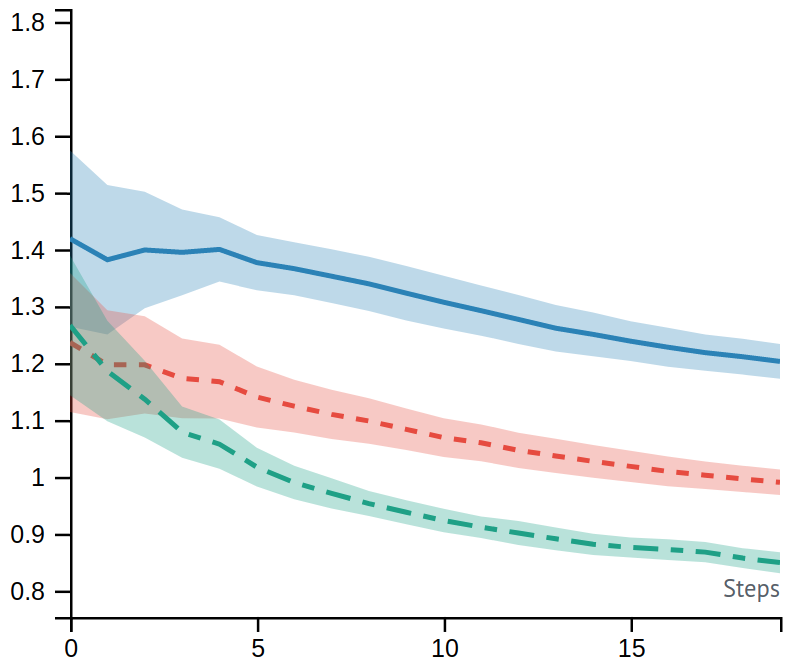}
         \caption{SuperLoss~\citep{castells2020superloss}}
         \label{fig:sl_conf}
     \end{subfigure}
     \quad
     \begin{subfigure}{0.47\linewidth}
         \centering
         \includegraphics[width=\linewidth]{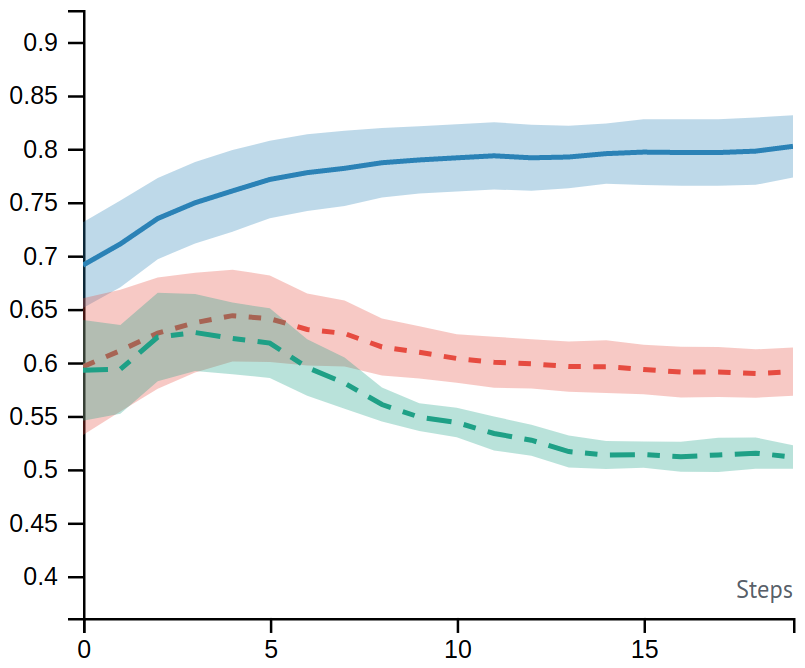}
         \caption{Self-paced Learning~\citep{kumar2010self}}
     \end{subfigure}
     \quad
     \begin{subfigure}{0.47\linewidth}
         \centering
         \includegraphics[width=\linewidth]{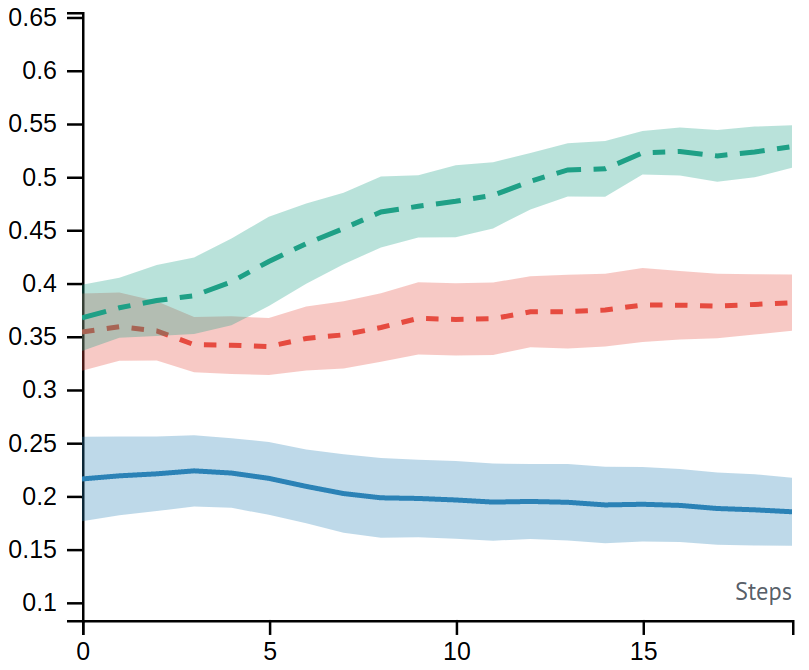}
         \caption{HNM~\citep{felzenszwalb2009object}}
     \end{subfigure}
    \caption{Confidence assignment to samples in our datasets by three CL approaches. The x-axis is the epoch number, and y-axis is the average weight assigned to samples of each difficulty group. Blue (solid) is \emph{easy}, orange (dashed) is \emph{medium}, and green (dash-dot) is \emph{hard}. The shaded area is the 95\% CI over the datasets with five random seeds each. The curves are monotonic for most parts, and can be approximated by our framework.}
    \label{fig:cl_curves}
\end{figure}

First, data pruning can be achieved by assigning negative values to the rate and shift parameters in our framework, $r$ and $s$ in (\ref{eq:weight}), which
cause the weights to approach zero before training begins. 
Second, data sub-sampling can be represented by ``inc'' in Figure~\ref{fig:inc_cfg}.
Third, approaches that estimate sample confidence based on loss~\citep{castells2020superloss,felzenszwalb2009object,kumar2010self,jiang2015self,zhou2020curriculum} tend to generate monotonic curves over the course of training because training loss tends to be non-increasing at every step. Figure~\ref{fig:cl_curves} shows the confidence scores assigned to our data by three loss re-weighting approaches. The results are generated by our implementations of the three approaches, where each model runs with five random seeds. The partitioning of {\em easy}, {\em medium}, and {\em hard} is according to the entropy, as described in \S\ref{sec:scoring}. We record the average weight assigned to each group. The result is averaged over all the runs, and the shaded area indicates the 95\% confidence interval (CI). The results show that the confidence scores assigned by these approaches follow a monotonic curve that can be approximated by our curriculum discovery framework. 
We note that although the weight scale of SuperLoss~\citep{castells2020superloss} in Figure~\ref{fig:sl_conf} is larger than one, this model can still be represented by our framework because the increased scale corresponds to scaling of the learning rate, as shown:
\begingroup
\setlength{\abovedisplayskip}{3pt}
\setlength{\belowdisplayskip}{3pt}
\begin{equation}
\begin{split}
    \theta_t & = \theta_{t-1} - \eta \nabla{\frac{1}{n} \sum_i{\sigma_i l_i}} \\
    & = \theta_{t-1} - (\eta \cdot \sigma_{max}) \nabla{\frac{1}{n} \sum_i{\frac{\sigma_i}{\sigma_{max}} l_i}},
\end{split}
\end{equation}
\endgroup
where $l_i$ and $\sigma_i$ are the instantaneous loss and confidence of sample $i$ respectively. Therefore, the proposed framework can also represent CL approaches with a confidence scale larger than one.

\section{Conclusion and Future Work}
\label{sec:conc}
We introduce an effective curriculum learning framework that employs prior knowledge about sample difficulty in its training paradigm for curriculum discovery. The proposed framework initially partitions its input data into several groups of increasing difficulty, defines parameterized functions to weight sample losses in each difficulty group, moves samples across difficulty groups based on their learning progress, and enables tuning the parameters of the weight function to discover novel curricula. We demonstrate that this framework is capable of representing several categories of curriculum learning approaches.
The task of curriculum discovery alleviates the limitations imposed by selecting a single curriculum strategy, and instead, focuses on finding and analyzing different curricula that work equally-well for a given model and dataset. In addition, the discovered curricula provide insight into how different portions of the dataset contribute toward learning at different stages of training a model, which, in turn, provide knowledge about the learning dynamics of different models. The task of curriculum discovery could be costly on large datasets, in particular, when the goal is to find optimal curricula for different models and datasets. To mitigate the computational cost, we show that it is possible to rapidly discover a curriculum on a small subset of the dataset (or a smaller version of the model with significantly less number of parameters) and apply the resulting curriculum to the full dataset. 



%

There are several promising areas for future work. These include approaches for learning new difficulty indicators from data (e.g., linguistic difficulty including lexical, syntactic and semantic difficulty), prioritizing medium level instances and those with greatest progress during training, and developing challenge datasets that contain diverse data samples with different levels of difficulty. Finally, investigating diverse curricula that are suitable for general use and across datasets through curriculum discovery and generalization is a promising area for research. 

\section*{Limitations}
The present work investigates the use of two sample difficulty scoring functions, human-induced annotation entropy and model-induced loss, for NLP models and datasets. The former requires the availability of multiple annotations per sample and the latter requires training an auxiliary model to compute sample instantaneous loss during the course of training. Our work does not provide a general solution to the choice or availability of good difficulty scoring functions.
However, once such a function is available, our work presents solutions to the problem of finding high-performing curricula in curriculum space. Our approach, although effective at finding such curricula, requires a Bayesian search of its hyperparameters. We reduce these costs by finding curricula on smaller datasets and smaller models that can then be applied to corresponding larger datasets and models. Finally, the proposed method 
lacks theoretical analysis of the dynamic interactions between data, downstream models, and discovered curricula.




\bibliography{custom}
\bibliographystyle{acl_natbib}

\onecolumn
\appendix

\section{Data Categories Distribution}
\label{sec:data_freq}
\begin{table*}[htb]\small
    \centering
    \begin{subtable}[htb]{.48\linewidth}
        \centering
        \begin{tabular}{l l }
        {\bf Class} & {\bf Count} \\
        \toprule
        (no)	& 5,325 \\
        (yes, light use, individual)	& 1,464 \\
        (yes, heavy use, individual)	& 964 \\
        (yes, not sure, individual)	& 457 \\
        (yes, heavy use, other)	& 423 \\
        (yes, heavy use, group) &	284 \\
        (yes, light use, group)	& 161 \\
        \midrule
        {\bf Total} & 9,078 \\
        \bottomrule
       \end{tabular}
       \caption{Twitter}
       \label{tab:alcohol_class}
    \end{subtable}
    \hfill
    \begin{subtable}[htb]{.48\linewidth}
        \centering
        \begin{tabular}{l l}
        {\bf Class} & {\bf Count} \\
        \toprule
        (irrelevant, no patient experience) &	1,996 \\
        (relevant, breast cancer) & 617 \\
        (relevant, colon cancer) &	444 \\
        (relevant, brain cancer) &	284 \\
        (irrelevant, none of the above) &	251 \\
        (irrelevant, other cancer types) &	162 \\
        (irrelevant, news related to cancer) &	70 \\
        \midrule
        {\bf Total} & 3,824 \\
        \bottomrule
        \end{tabular}
        \caption{Reddit}
        \label{tab:cancer_class}
     \end{subtable}
     \caption{Statistics of the Twitter and Reddit datasets.}
     \label{tab:data_class}
\end{table*}
Table~\ref{tab:data_class} shows the target class distributions of the Reddit and Twitter datasets.

\section{Finer-grained Difficulty Classes}
\label{sec:fg}
\begin{figure}[htb]
     \centering
     \begin{subfigure}[htb]{0.23\linewidth}
         \centering
        \includegraphics[width=\linewidth]{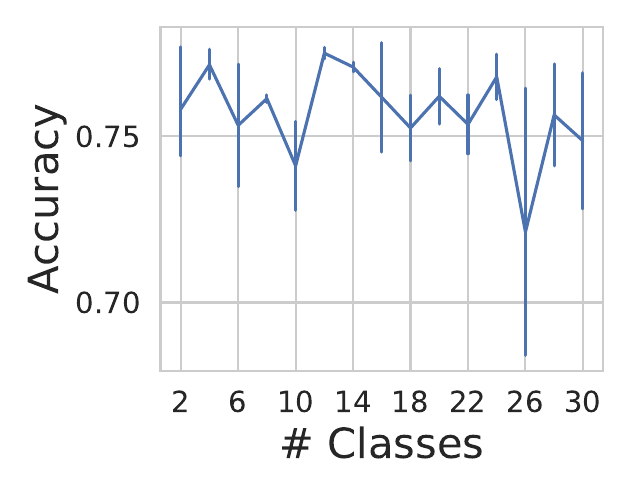}
        \caption{ChaosNLI.}
     \end{subfigure}
     \begin{subfigure}[htb]{0.23\linewidth}
         \centering
        \includegraphics[width=\linewidth]{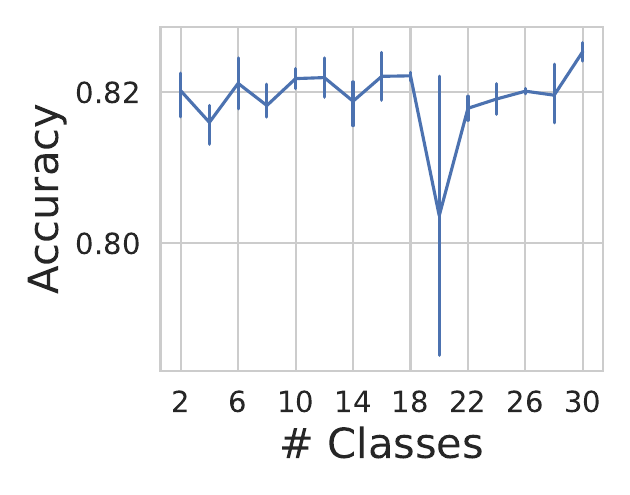}
        \caption{SNLI.}
     \end{subfigure}
     \begin{subfigure}[htb]{0.23\linewidth}
         \centering
        \includegraphics[width=\linewidth]{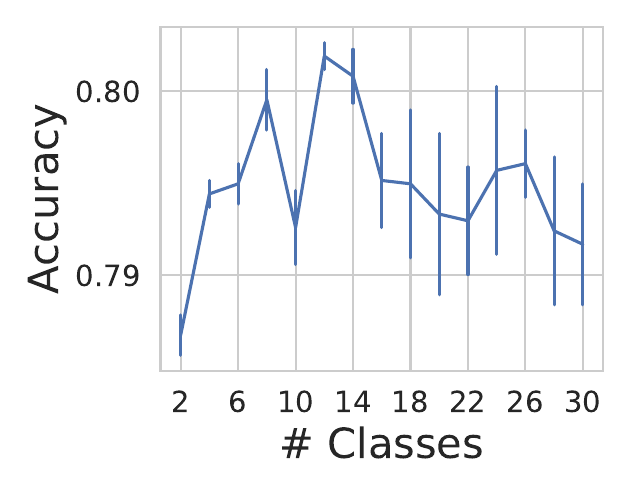}
        \caption{Twitter.}
     \end{subfigure}
     \begin{subfigure}[htb]{0.23\linewidth}
         \centering
        \includegraphics[width=\linewidth]{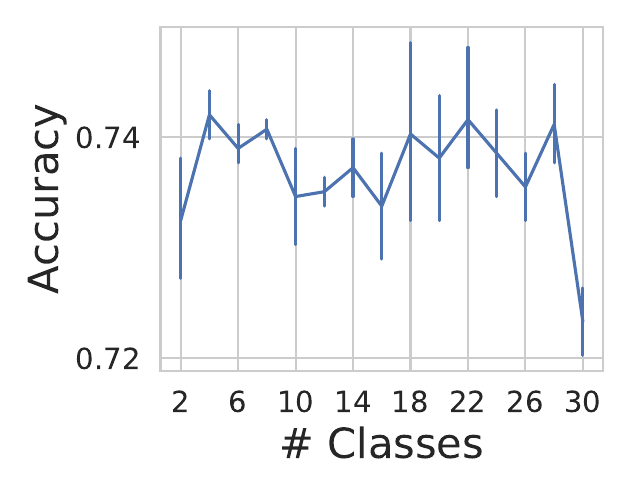}
        \caption{Reddit.}
     \end{subfigure}
    
    \caption{Accuracy of models trained with the inc curriculum (see \S\ref{sec:baselines}) and different number of difficulty classes.}
    \label{fig:fg}
\end{figure}

Figure~\ref{fig:fg} shows the effect of different number of difficulty classes on he accuracy of models trained with our \textit{inc} curriculum (see \S\ref{sec:baselines}). The results show that the number of difficulty classes used is an important factor in our framework, and further tuning of this parameter can further improve the performance of our model.  

\section{Curriculum Search Computational Cost}
\label{sec:computation}
    \begin{table}[htb]\small
        \centering
        \begin{tabular}{l c}
        {\bf Configuration} & {\bf Number of trials} \\
        & {\small (Avg. turnaround time per trial: 15 minutes)} \\ \toprule
        S-F-E & 87 \\
        S-F-L & 111 \\
        S-B-E & 135 \\
        S-B-L & 75 \\
        T-F-E & 139 \\
        T-F-L & 73 \\
        T-B-E & 106 \\
        T-B-L & 44 \\
        R-F-E & 61 \\
        R-F-L & 73 \\
        R-B-E & 69 \\
        R-B-L & 112 \\
        C-D-E & 36 \\
        C-D-L & 70 \\
        C-D-E [82M parameter model] & 71 \\
        C-D-E [406M parameter model] & 69 \\
        \bottomrule
        \end{tabular}
        \caption{Number of trials for the best parameters found. The notation for configurations is the same as Figure~\ref{fig:configs}.}
        \label{tab:computation}
     \end{table}

With our experimental settings, it takes around 15 minutes on average to train a base model on our datasets of up to 3k samples using a single GPU. Therefore, a curriculum search take around 9 hours (36 trials) to around 35 hours (139 trials) using a single GPU.

\section{Extended Configuration Generalizablity Experiments}
\label{sec:nxn_full}
\begin{figure}[htb]
    \centering
    \includegraphics[width=0.8\linewidth]{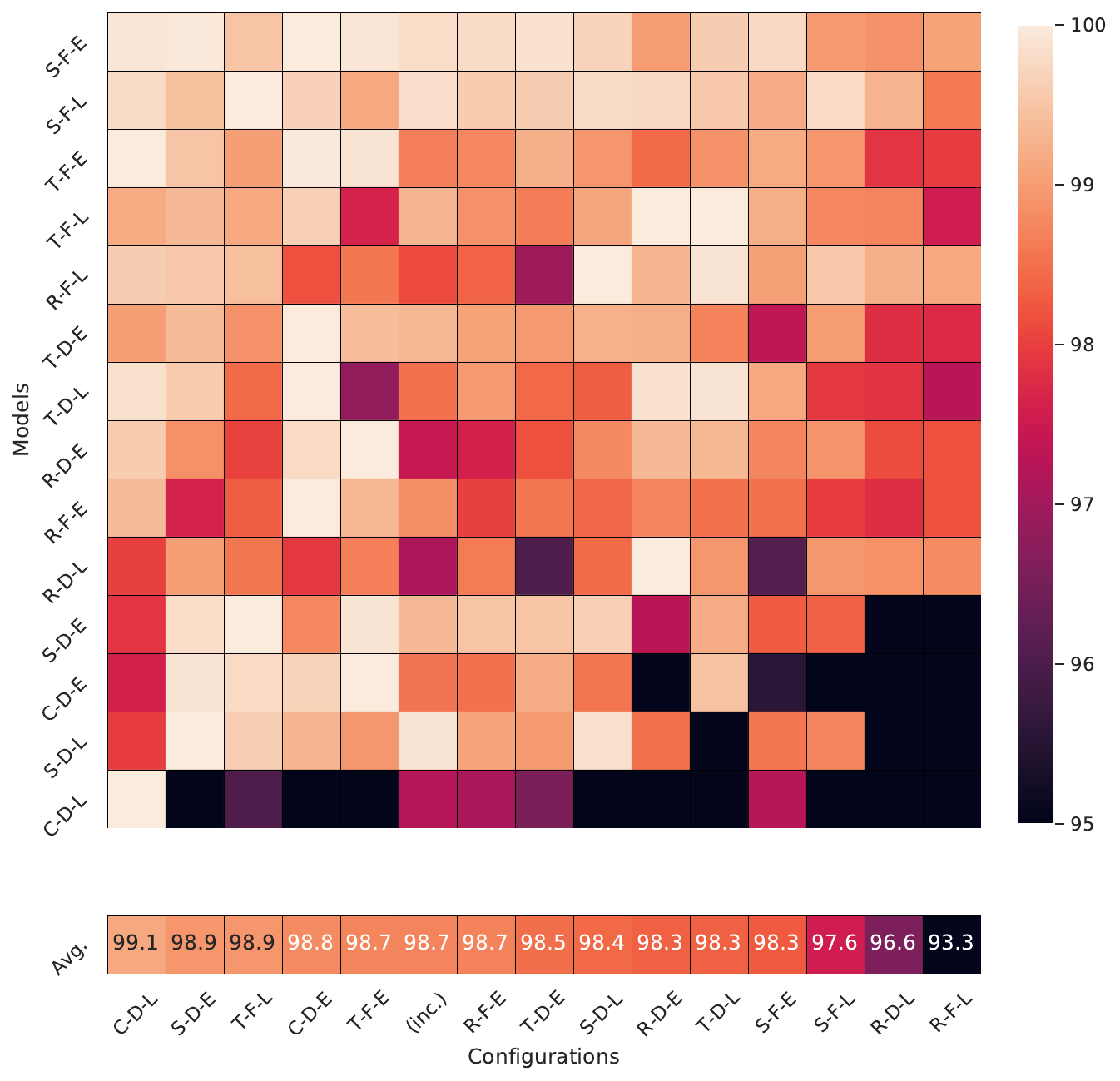}
    \caption{An extended version of Figure~\ref{fig:nxn} including experiments on balanced versions of the datasets.}
    \label{fig:nxn_full}
\end{figure}
Figure~\ref{fig:nxn_full} shows the result of every model trained using every specialized curricula (and {\em inc}). We see that the generalizable curricula that are effective on small (down-sampled) datasets, also tend to perform well on large (full) datasets.

\end{document}